\documentclass[10pt,twocolumn,letterpaper]{article}

\usepackage{cvpr}
\usepackage{times}
\usepackage{epsfig}
\usepackage{graphicx}
\usepackage{amsmath}
\usepackage{amssymb}

\newcommand{\eat}[1]{}

\cvprfinalcopy % *** Uncomment this line for the final submission

\def\cvprPaperID{69} % *** Enter the CVPR Paper ID here

\ifcvprfinal\fi
\begin{document}

%%%%%%%%% TITLE
\eat{
\title{Simple baselines for Video Super-Resolution: Adapting Image Super-Resolution State-of-the-arts and Learning Multi-model Ensemble}

}
\title{Adapting Image Super-Resolution State-of-the-arts and Learning Multi-model Ensemble for Video Super-Resolution}

\author{Chao Li, Dongliang He, Xiao Liu, Yukang Ding, Shilei Wen\\
Department of Computer Vision Technology (VIS), Baidu Inc.\\
{\tt\small \{lichao40,hedongliang01,liuxiao12,dingyukang,wenshilei\}@baidu.com}
% For a paper whose authors are all at the same institution,
% omit the following lines up until the closing ``}''.
% Additional authors and addresses can be added with ``\and'',
% just like the second author.
% To save space, use either the email address or home page, not both
}

\maketitle
%\thispagestyle{empty}

%%%%%%%%% ABSTRACT
\begin{abstract}
    Recently, image super-resolution has been widely studied and achieved significant progress by leveraging the power of deep convolutional neural networks. However, there has been limited advancement in video super-resolution (VSR) due to the complex temporal patterns in videos. In this paper, we investigate how to adapt state-of-the-art methods of image super-resolution for video super-resolution. The proposed adapting method is straightforward. The information  among successive frames is well exploited, while the overhead on the original image super-resolution method is negligible. Furthermore, we propose a learning-based method to ensemble the outputs from multiple super-resolution models. Our methods show superior performance and rank second in the NTIRE2019 Video Super-Resolution Challenge Track 1.
\end{abstract}

%%%%%%%%% BODY TEXT
\section{Introduction}

\begin{figure*}[h]
\begin{center}
\includegraphics[width=\linewidth]{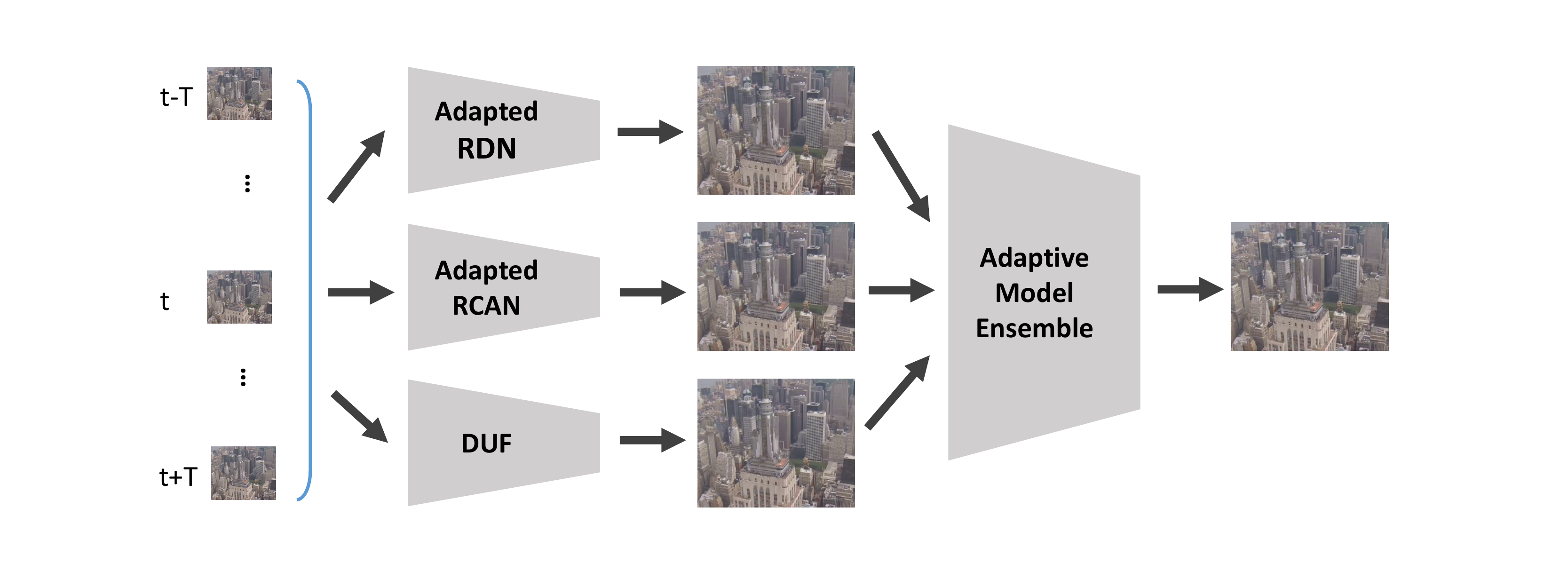}
%\fbox{\rule{0pt}{2in} \rule{0.9\linewidth}{0pt}}
   %\includegraphics[width=0.8\linewidth]{egfigure.eps}
\end{center}
   \caption{Demonstration of the Workflow of our framework for VSR. We adapt state-of-the-art image super-resolution frameworks, i.e., RCAN~\cite{eccv/Zhang18} and RDN~\cite{cvpr/Zhang18} for video super-resolution. Besides, the 3D convolution based method DUF~\cite{/cvpr/Jo18} is also used as one of our baseline model. The results generated by these models are ensembled via a novel adaptive model ensemble module to produce the final result.}
\label{fig:workflow}
\end{figure*}

Image super-resolution (ISR) has drawn extensive attention in recent decades. Taking a low-resolution image as input, ISR aims to generate a high-resolution image with more visual details. It has many applications in various fields, such as security surveillance, medical imaging. By exploiting deep convolutional neural networks, ISR has gained remarkable success. 
Dong et al. proposed a three-layer convolutional neural network for ISR and achieved significant improvement over conventional methods~\cite{iccv/Timofte13,tip/Zhang06,tip/Zhang12}. Kim et al. developed a much deeper network in VDSR~\cite{cvpr/Kim16a} and DRCN~\cite{cvpr/Kim16b} by using gradient clipping, skip connection, or recursive-supervision to handle the difficulty of training deep network. 
Lim et al. employed enhanced residual blocks and residual scaling to build a very wide network EDSR~\cite{cvpr/Lim17}.
In~\cite{cvpr/Zhang18} a residual dense network (RDN) was proposed to fully utilize the all the hierarchical features by leveraging dense connection and residual learning. 
In~\cite{eccv/Zhang18}, Zhang et al. designed a residual channel attention network (RCAN) to adaptively rescale channel-wise features by considering interdependencies among channels.

Although the performance of image super-resolution has been significantly advanced by these above techniques, video super-resolution still needs to be promoted because of the high demand for VSR in real applications (e.g. video on demand services and live-broadcasting platforms) and its unsatisfactory performance with regard to visual quality and computation complexity. For video super-resolution, to exploit the plentiful temporal information among successive frames, several methods has been proposed.
Tao et al.~\cite{/iccv/Tao17} used motion compensation transformer module for the motion estimation, and
proposed a sub-pixel motion compensation layer for simultaneous motion compensation and upsampling.
Sajjadi et al.~\cite{/cvpr/Sajjadi18} extended the conventional VSR model to a frame-recurrent VSR framework by using the previously inferred high-resolution frame to super-resolve the subsequent frame.
The above methods heavily rely on the accuracy of flow estimation and motion compensation which also bring extra computation cost. 
Jo et al.~\cite{/cvpr/Jo18} introduced a fundamentally different framework for VSR. They proposed an end-to-end deep neural network that generates dynamic upsampling filters and a residual image, which were computed depending on the local spatio-temporal neighborhood of each pixel to avoid explicit motion compensation. However, they employed 3D convolution kernels to model the temporal information among successive frames, which are slow to compute comparing to using 2D convolution kernels.

To address the aforementioned issues, we propose to adapt state-of-the-art ISR methods for VSR while keep the computation cost staying in the original level. A demonstration of our framework for VSR is in Figure~\ref{fig:workflow}. Firstly, we optimize state-of-the-art ISR methods by analyzing and modifying their network components to enable them to handle VSR task. The adapted methods significantly outperform the original methods which process a video frame-by-frame. Secondly, we propose a learning-based method to ensemble the outputs from multiple super-resolution models which again boosts the final performance remarkably. Leveraging the adapted super-resolution methods and the learning-based ensemble method, we ranked second in the NTIRE2019 Video Super-Resolution Challenge Track 1~\cite{Nah_2019_CVPR_Workshops_SR}.

\eat{
With our approach, an HR image is reconstructed directly from the input image using the dynamic upsampling filters, and the fine details are added through the computed residual.

Recent research on VSR
flow hard to train

By using effective building modules, the networks for image SR are further made deeper and wider with better performance. 

EDSR, 
RDN, 
RCAN,

video sr
frame by frame, information lost

conv lstm 
flow
detail revealing iccv

Our method 
adapting 
conv ensemble
}

%-------------------------------------------------------------------------

%------------------------------------------------------------------------
\section{Related Work}
\subsection{Image Super-resolution}

Conventional image super-resolution methods use interpolation techniques based on sampling theory~\cite{tip/LiO01,tip/Zhang06},  and several previous studies~\cite{cvpr/Tai10,cvpr/Sun08} adopted natural image statistics to restore high-resolution images. However, these methods can not reconstruct satisfactory details and realistic textures.
By utilizing techniques, such as neighbor embedding and sparse coding~\cite{bmvc/Bevilacqua12,cvpr/Chang04,tip/Gao12,tip/Yang10,accv/Timofte14}, some learning-based methods attempt to learn mapping functions between low-resolution and high-resolution image pairs.

Recently, with the development of deep convolutional neural networks (CNN), ISR gains dramatic improvements. Dong et al.~\cite{eccv/Dong14,eccv/Dong16} propose a CNN-based super-resolution method. Afterwards, various CNN architectures have been investigated for ISR. Kim et al.~\cite{cvpr/Kim16a,cvpr/Kim16b} employs the residual network for training much deeper network and achieved superior performance. 
In particular, they use skip-connection and recursive convolution to alleviate the burden of carrying identity information in the super-resolution network.
In ~\cite{nips/Mao16} Mao et al. utilize encoder-decoder networks and symmetric skip connections~\cite{miccai/Ronneberger15} to achieve fast and improved convergence. Futhermore, Lim et al. employed enhanced residual blocks and residual scaling to build a very wide network EDSR~\cite{cvpr/Lim17}.
In~\cite{cvpr/Zhang18} a residual dense network (RDN) was proposed to fully utilize the all the hierarchical features by leveraging dense connection and residual learning. 
In~\cite{eccv/Zhang18}, Zhang et al. designed a residual channel attention network (RCAN) to adaptively rescale channel-wise features by considering interdependencies among channels. 
DBPN~\cite{cvpr/Haris18} exploits iterative up and down sampling layers to build an error feedback mechanism for errors projection at each stage. The authors construct mutually connected up and down sampling stages each of which represents different types of image degradation and high-resolution components.
%-------------------------------------------------------------------------
\subsection{Video Super-resolution}

For video super-resolution, previous methods usually adopted two-stage
approaches that are based on optical flow. In the first stage, motion estimation is conducted by computing optical flow. In the second stage, the estimated motion fields are used to perform image wrapping and motion compensation. For example, Liao et al.~\cite{iccv/Liao15} used classical optical flow methods to generate high-resolution image drafts, and then predicted the final high-resolution frame by a deep draft-ensemble network. 
However, the classical optical flow algorithms are independent of the frame reconstruction CNN and much slower than the flow CNN during inference.
To tackle this issue, Caballero et al.~\cite{cvpr/Caballero17} design an end-to-end VSR network to jointly trains flow estimation and spatio-temporal networks.
Tao et al.~\cite{/iccv/Tao17} compute low-resolution motion field based on optical flow network and develop a new layer to utilize sub-pixel information from motion and achieve sub-pixel motion compensation (SPMC) and resolution enhancement simultaneously. 
Xue et al.~\cite{corr/Xue17} exploit task-oriented flow to get better VSR results than fixed flow algorithms.
Sajjadi et al.~\cite{/cvpr/Sajjadi18} propose a frame-recurrent VSR framework which uses the previously  inferred high-resolution frame to super-resolve the subsequent frame.
However, it is non-trivial to obtain high-quality motion estimation even with state-of-the-art optical flow estimation networks. 
Even with accurate motion fields, the image-wrapping based motion compensation will bring artifacts to super-resolved images, which could be propagated into the reconstructed high-resolution frames.
Instead of explicitly computing and compensating motions between input frames, Jo et al.~\cite{/cvpr/Jo18} proposed to utilize the motion information implicitly via generating dynamic upsampling filters. With the generated upsampling filters, the high-resolution frame is constructed by local filtering to the input center
frame. However, they employ 3D convolution kernels to model the temporal information among successive  frames, which are slow to compute comparing to using 2D convolution kernels.

%-------------------------------------------------------------------------
\section{The Proposed Framework}

\begin{figure*}[h]
\begin{center}
\includegraphics[width=0.8\linewidth]{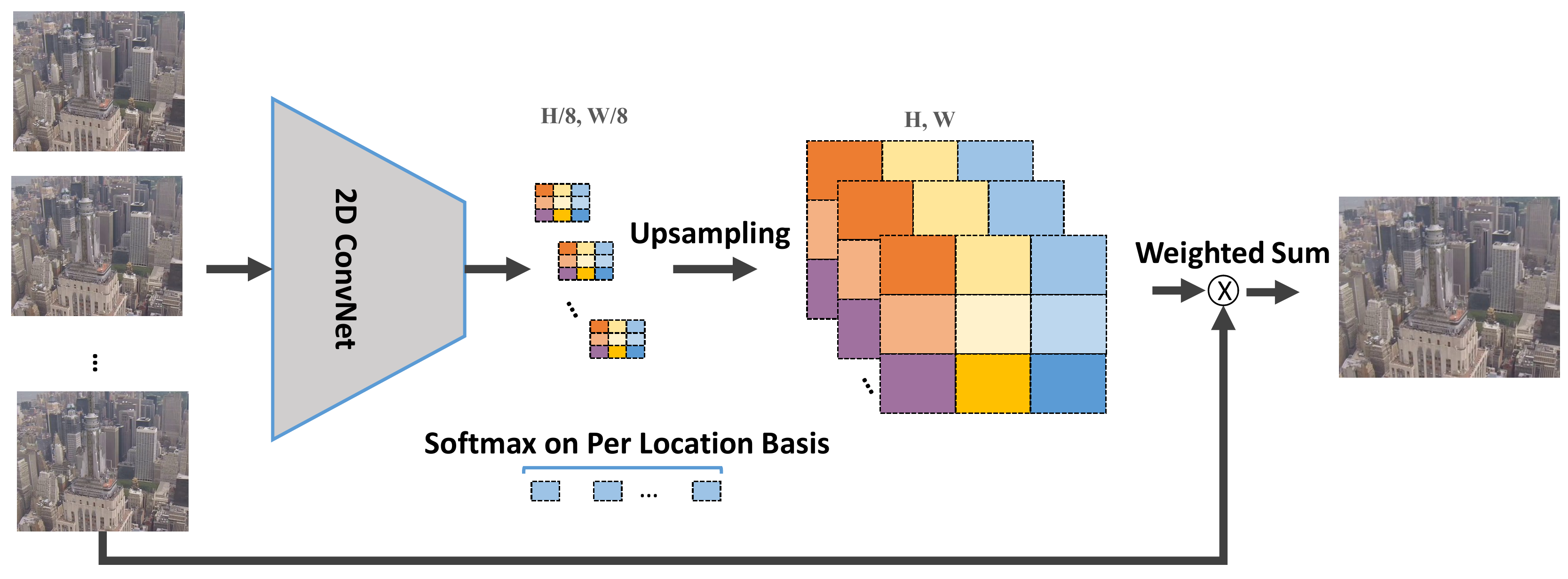}
\end{center}
   \caption{Illustration of the proposed adaptive model ensemble framework. The inputs are the frames generated by $N$ models. A 2D ConvNet with stride of 8 is employed to extract a $1\times H/8\times W/8$ feature map for each image. Then, a softmax activation is applied to each spatial location of these feature maps, and afterwards the activation score map is up-sampled by a factor of 8. In this way, the fusion weights of all models sum to 1 at every pixel. }
\label{fig:ensemble}
\end{figure*}

In this section, we describe our method for video super-resolution. The task is to map low resolution frames of a video into up-sampled high resolution ones. Different from image super-resolution, video frames not only demonstrate spatial correlation in each frame, but also contain temporal information among neighbouring frames. How to effectively and efficiently learn spatial-temporal feature to produce a super-resolved video frame is the key to VSR. Generally speaking, we propose to adapt multiple state-of-the-art image super-resolution methods for video super-resolution. Furthermore, for model ensemble, a CNN model is designed to adaptively ensemble results from multiple models.

\subsection{Adapting Image Super-resolution State-of-the-arts for Video Super-resolution}
We propose to learn deep spatial-temporal features for up-sampling video frames by adapting multiple state-of-the-art image super-resolution methods. Our insight is that, image super-resolution methods, which use 2D convolutional kernels to capture spatial features from images, can efficiently be adopted for spatial-temporal feature modeling by applying 2D convolutional filtering on an ``super image'' consists of several successive video frames. The image super-resolution methods adapted by us for video super-resolution including RCAN \cite{eccv/Zhang18} and RDN \cite{cvpr/Zhang18}. To up-sample the $t$-th frame $F_t$, firstly, we early fuse successive frames of $[F_{t-T}, ..., F_{t-1}, F_t, F_{t+1}, ..., F_{t+T}]$ by concatenating them of along the channel dimension, such that the input ``super-image'' is a $3(2T+1)\times H_{lr} \times W_{lr}$ tensor, where $H_{lr}$ and $W_{lr}$ denote the height and width, respectively. Then the first convolution layer of RDN or RCAN is manipulated to accept ``super-image'' as input by changing its input channel number from 3 to $3(2T+1)$. The advantage of this adaptation schema is to largely reduce the computation cost compared with using 3D based solutions, such as DUF proposed in \cite{/cvpr/Jo18}, and its effectiveness is also verified by our experimental results.

\subsection{Learning Multi-model Ensemble}

We design a learning based adaptive ensemble method to leverage the diversity among different models. We notice that even with the same model, in a super-resolved frame, the visual quality of different patches vary a lot. These priors motivate us to ensemble multiple models in an adaptive way, namely, the fusion weights of different models must be conditioned on the frames generated by these models in an image patch granularity. 

The framework of our adaptive model ensemble solution is depicted in Figure~\ref{fig:ensemble}. We use a 2D ConvNet to model the patch-level diversities among different models. The 2D ConvNet contains 3 Conv-BN-ReLU layers, the number of output channels of the 3 layers are 16, 32 and 64 respectively. The kernel size and the stride of all the 3 convolution layers are both set to 2, and no padding is used in this model. After the 3 Conv-BN-ReLU layers, a 1x1 convolution layer with output channel size of 1 is attached to generate a $1\times \frac{H}{8}\times \frac{W}{8}$ score map for each input image. To ensemble images of $N$ models needs to produce $N$ score maps, we apply softmax to each spatial location of the score maps along the $N$ model dimension and then $8\times$ up-sampling is used. Therefore, every non-overlapped $8\times8$ patch shares a same score and the scores of the $N$ maps sum to 1 at every spatial location. These final score maps are treated as fusion weights for each model. We can see that the proposed ensemble method can be trained to adaptively ensemble the results of multiple models at $8\times8$ patch granularity yet at very limited cost.  

\section{Experiments}

\subsection{Dataset}
\label{sec:dataset}
We conduct experiments on a recently released novel dataset of low and high resolution blur-free videos (REDS~\cite{Nah_2019_CVPR_Workshops_REDS}) obtained in diverse environments. It enables an accurate benchmarking of the performances achieved by the video super-resolution methods.
In the REDS dataset, there are 300 pairs of low and corresponding high resolution blur-free video sequences and each video sequence is 100 frames long. For the NTIRE19 competition, REDS is divided into 240 video pairs for training and 30 for validation and 30 for testing. 

The groundtruth of the 30 testing videos are not publicly available, therefore our experiments are performed on the training and validation videos. During the development phase of the NTIRE2019 competition, we split the 240 training video pairs into two parts, i.e., 216 for training and 24 (Val24) for model selection, and the 30 validation videos (Val30) are used for testing the performance of our model. During the final phase of the NTIRE2019 competition, besides the 216 training videos and 30 validation videos, we further split 12 video from the Val24 set for training, finally we got 258 videos for model training and 12 videos (Val12) for model selection. 

\subsection{Implementation Details}
~\label{sec:impl}
We investigate and adapt three ISR state-of-the-arts, i.e., EDSR~\cite{cvpr/Lim17}, RDN~\cite{cvpr/Zhang18}, RCAN~\cite{eccv/Zhang18} for VSR. For each method, we adopt the configuration corresponding to the best performance reported in the orginal papers.
L1-Loss is used to train the EDSR, RDN and RCAN models, and For DBPN, MSE loss is used. We employ ADAM optimizer and set initial learning rate to $1e-4$. We decrease the learning rate to $5e-5$, $3e-5$ and $1e-5$ sequentially when it meets a converge point. 
To validate the effectiveness of our ensemble method. We also trained a DUF model with L1-Loss and SGD, its learning rate is initially set to 0.1 and is $10\times$ decreased every 50 passes. 
The adaptive ensemble CNN model are trained on the Val12 set which is reserved by us for model selection. Its learning strategy is the same as DUF. All the models are trained from scratch.
Four kinds of augmentation are employed, i.e. 90 degree rotation, vertical flip, horizontal flip and temporal flip. For a training sample, each augmentation is adopted with probability of 0.5. In the training phase, $96\times96$ patches are random cropped. 

For testing, following~\cite{cvpr/Lim17}, we use self-ensemble to test each single model. In our setting, there are 16 augmentations in total by combining temporal flipping, vertical flip, horizontal flip and spatial rotation. In testing, the whole low-resolution ``super image'' is fed into each single model to produce up-sampling result with $16\times$ self-ensemble. 
Temporal reflection padding is utilized for the $t$-th input ``super-image'' if any element in [$t-T$,...,$t-1$, $t$, $t+1$,..., $t+T$] is out of range, $t=0,1,...,99$
The proposed adaptive ensemble model is employed to ensemble the results from each single model. The PSNR metric is used for performance evaluation. 

\subsection{Evaluation of Adapted Methods}
\begin{table}
\begin{center}
\begin{tabular}{|c|c|c|}
\hline
Method & Image SR & Adapted Video SR \\
\hline\hline
EDSR &28.52  & 29.55 \\
RCAN &28.63  & 29.93 \\
RDN  &28.52  & 29.71 \\ \hline
\end{tabular}
\end{center}
\caption{Comparison of original Image SR methods and the corresponding adapted methods for Video SR on Val24. All these models are trained on 216 videos and tested on Val24. No self-ensemble is used.}
\label{tab:frame-ezconv}
\end{table}

\begin{figure*}[h]
\begin{center}
\includegraphics[width=0.76\linewidth]{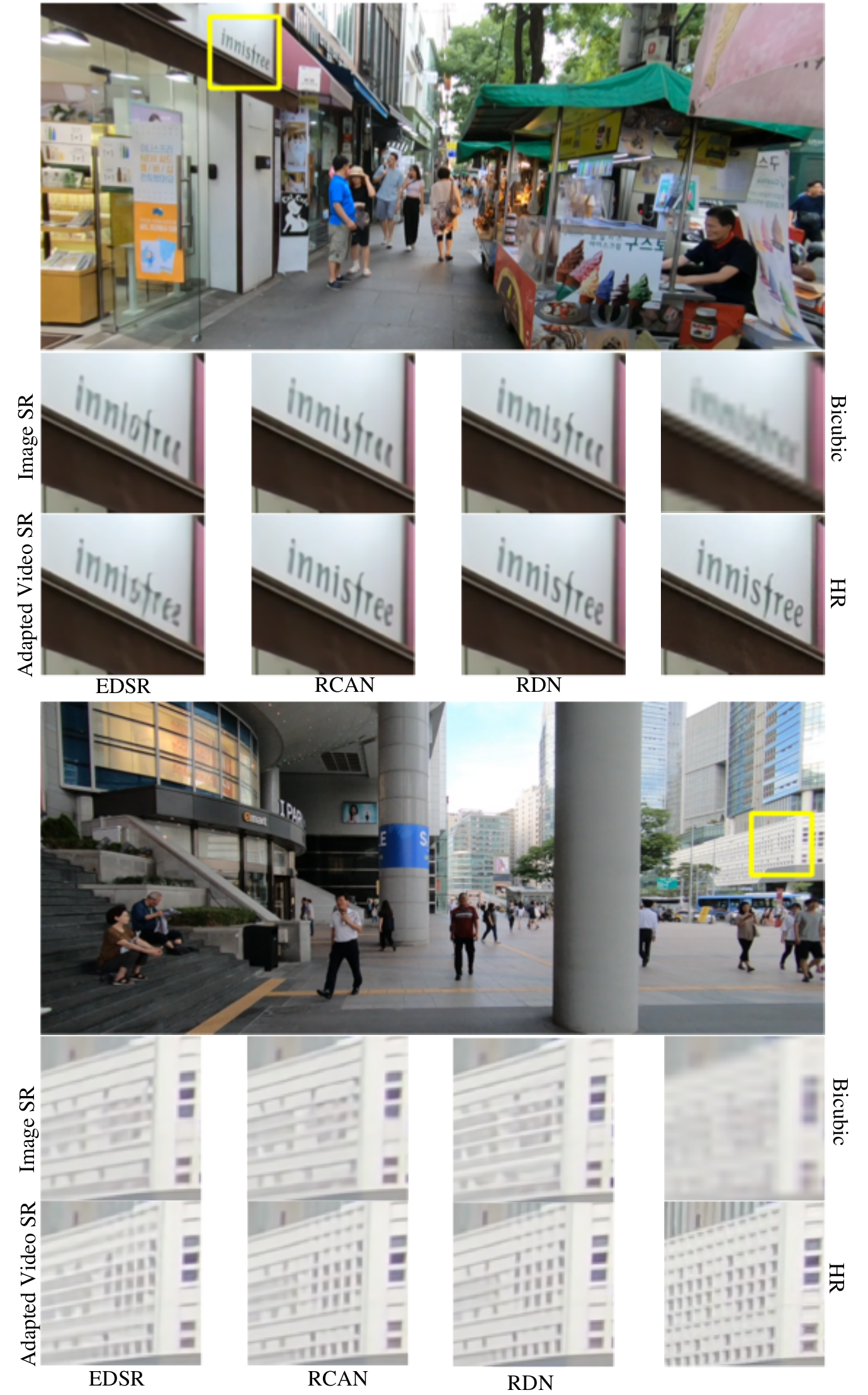}
%\fbox{\rule{0pt}{2in} \rule{0.9\linewidth}{0pt}}
   %\includegraphics[width=0.8\linewidth]{egfigure.eps}
\end{center}
   \caption{Qualitative comparison of our adapted Video SR models with the corresponding original Image SR models.}
\label{fig:demo}
\end{figure*}

To evaluated the proposed adapting method, we train the original ISR methods, i.e., EDSR~\cite{cvpr/Lim17}, RDN~\cite{cvpr/Zhang18}, RCAN~\cite{eccv/Zhang18}, and their corresponding adapted variants, on the 216 training videos. The evaluation results on Val24 set is reported in Table~\ref{tab:frame-ezconv}, and no self-ensemble is used for each method in testing. The length of the input ``super image'' for each model is set to 5, i.e., the $T$ as mentioned in Section~\ref{sec:impl} is set to 2. As shown in the table, All the three adapted VSR methods outperform their original ISR version significantly. It proves the effectiveness of the proposed adapting method for VSR. We also provide qualitative results in Figure~\ref{fig:demo}. The adapted models successfully reconstruct the detailed textures and edges in the high-resolution images and present super-resolved outputs with better visual quality than the original ISR methods.

\subsection{Learning Multi-model Ensemble}

\begin{table}
\begin{center}
\begin{tabular}{|c|c|c|}
\hline
Method & Val30 & Val12 \\
\hline\hline
RCAN &30.13  & 29.97 \\
RDN &30.14  & 30.01 \\
RDN-MSE &30.11  & 29.94 \\ 
RDN-Deeper & $\times$ &30.03 \\
RDN-Extradata &$\times$ &30.03 \\
RDN-Bicubic  &$\times$  &29.91 \\
DUF &29.78 &29.58 \\ \hline\hline
Avg Ensemble &30.20 &30.05 \\
Adaptive Ensemble &30.26 &30.07 \\
\hline
\end{tabular}
\end{center}
\caption{Quantitative Results of different methods in terms of PSNR in dB. Val30 is the official validation set, on which the results in the developing phase are reported. Val12 is our own division on which the results trained on 258 videos in the final phase are reported. $\times$ means not used.}
\label{tab:result}
\end{table}

In the final phase of the NTIRE2019 VSR competition, five adatped RDN variants are trained, i.e., the adatped RDN network with L1 Loss (RDN), the adatped RDN network with MSE loss (RDN-MSE), the adapted RDN network with more dense blocks with L1 Loss (RDN-Deeper), the adapted RDN network augmented by adding a bicubic $4\times$ up-sampled residual connection (RDN-Bicubic) and the adapted RDN trained with extra data (RDN-Extradata). All the above RDN model are configured with the best setting in the original paper~~\cite{cvpr/Zhang18}, except that we increase the channel number from 64 to 128. For the RCAN~\cite{eccv/Zhang18} model, we find that enlarge the channel numbers gains no performance improvement. We analyze the reason is that, RCAN utilizes channel attention mechanism which effectively exploits the information among each channel and the original channel number is already enough for fitting the REDS dataset. However, we still improve the performance by adding more blocks to RCAN. To validate the effectiveness of our ensemble method, besides the above adapted VSR methods, we also trained a DUF model as described in Section~\ref{sec:impl}.   
To increase the model diversity, we use different $T$ settings for the above models. The $T$ settings of RCAN, RDN, RDN-MSE, RDN-Deeper, RDN-Extradat, RDN-Bicubic and DUF are 2,3,1,3,3,2 and 2 respectively.

For training the RDN-Extradata model, we manually select 9 videos from Youtube website and 30 video clips, each of which with length of 4 seconds, are randomly trimmed from them. Then we downscale the video clips with a factor of 4 and decode each of them into 100 frames. The video ids of the 9 videos are \emph{44ZIHUIybM8, 9AX1\_eEmRMU, AK80adiQVPw, dpgHqCrCmwY, ITpBDhc6p-0, L5FKHsFQ0IA, mjbUEuZtZ08, OU5ipJnApOY and zW\_UqsYuhbY}. These extra data are used for finetuning our models in the test phase. 

We report the experimental results in Table \ref{tab:result}. The results on Val30 are from models trained on the 216 training videos and the results on Val12 are produced by models trained on the 258 traing videos, as mentioned in Section~\ref{sec:dataset}
As shown in the table, the learned adaptive ensemble outperforms the average ensemble by a remarkable margin, which evidences the effectiveness of the proposed ensemble method.
The proposed adaptive ensemble method is data driven and can be trained to adaptively assign fusion weight to each single model. Compared to single method, such fusion method can better leverage model diversity to boost the performance.

\section{Conclusion}
In this paper we propose to adapt image super-resolution state-of-the-arts for video super-resolution. The proposed adapting method is straightforward to implement and the extra adapting overhead is negligible. Different from existing video super-resolution methods, our framework does not rely on flow estimation nor motion compensation, which are slow to compute and may propagate errors to high-resolution image reconstruction. Besides, our method only involves 2D convolution kernels rather than 3D convolution kernels as in DUF~\cite{/cvpr/Jo18}. Compared to 3D Convolution, our method is also an efficient way to learn deep spatial-temporal features from videos. Therefore, our adapting method can handle video super-resolution task while the computation cost is the same as image super-resolution. The experimental results on the recent benchmark dataset for video super resolution, i.e., REDS, have verified the effectiveness of the proposed adapting method.  

Furthermore, we propose an adaptive model ensemble framework, which can effectively exploit the diversity of the results from every single model. It is lightweight and can boost the performance remarkably. The experimental results on REDS also evidences its effectivenee.

%-------------------------------------------------------------------------

{\small
\bibliographystyle{ieee_fullname}

}

\end{document}